\documentclass[letterpaper, 10 pt, conference]{ieeeconf}  

\IEEEoverridecommandlockouts                              

\usepackage{graphics} 
\usepackage{amsmath} 
\usepackage{amssymb}  

\usepackage{booktabs}
\usepackage{multirow}
\usepackage{graphicx}
\usepackage{subcaption}  
\usepackage{censor}

\title{\LARGE \bf
DAP-Pose: Deep Temporal Alignment and Physics-aware Cross-modal Sensor Fusion for Robust Pose Estimation
}

\author{Jianhan Lin, Yuchu Qin, Jiateng Yuan, Wenbo Zhang and Shuai Gao 
\thanks{Jianhan Lin and Jiateng Yuan are with Aerospace Information Research Institute, Chinese Academy of Sciences, Beijing 100094, China. International Research Center of Big Data for Sustainable Development Goals, Beijing 100094, China. University of Chinese Academy of Sciences, School of Electronic, Electrical and Communication Engineering, Beijing 100049, China.}%
\thanks{Yuchu Qin and Shuai Gao are with International Research Center of Big Data for Sustainable Development Goals, Beijing 100094, China. Aerospace Information Research Institute, Chinese Academy of Sciences, Beijing 100094, China (e-mail:qinyc@aircas.ac.cn)(Corresponding author:Yuchu Qin).}
\thanks{Wenbo Zhang is with The University of Adelaide, Adelaide, SA 5005, Australia.}
}

\begin{document}

\maketitle
\thispagestyle{empty}
\pagestyle{empty}

\begin{abstract}
Robust and accurate pose estimation with multi-modal sensors is fundamental for autonomous vehicles and mobile robotic systems in complex environments. In this paper, we propose DAP-Pose, a unified end-to-end model for robust multi-modal pose estimation. DAP-Pose introduces a Bi-level Cross-modal Fusion (BCF) module that captures complementary semantic and geometric motion cues from visual, inertial, and GNSS measurements. To handle temporal offsets, we designed a Deep Temporal Alignment (DTA) module that explicitly aligns asynchronous streams in latent space, enabling coherent motion modeling without strict hardware synchronization. Furthermore, we incorporate physics-aware constraints via manifold geometry and GNSS-guided absolute metric scale, enforcing motion consistency and mitigating drift. Experiments upon the public KITTI benchmark dataset were conducted to evaluate the performance of DAP-Pose against existing methods. DAP-Pose achieved the state-of-the-art performance, with the lowest average translation error ($t_{rel}$) of 1.31\% and rotation error ($r_{rel}$) of 0.46$^{\circ}$. Furthermore, it accurately estimates poses and maintains robust performance under severe artificially injected temporal misalignment.

\end{abstract}

\section{INTRODUCTION}
Pose estimation, i.e., localization and orientation identification, for autonomous vehicles and mobile systems relies on heterogeneous data sets acquired by various sensors, e.g. camera, LiDAR, GNSS, and Inertial Measurement Unit (IMU). Robust and accurate pose estimation is essential for the safe and efficient path planning and navigation \cite{bib1}. Therefore, fusion of multi-sensor observations was widely adopted for localization and estimation of orientation.

Traditional filter-based and optimization-based algorithms, such as Kalman filtering and factor graph optimization, were widely applied in pose estimation. However, those methods largely depend on the accurate modeling of system dynamics and predefined noise distributions, which fundamentally limits their robustness under severe nonlinearities. Simultaneous Localization and Mapping (SLAM) approaches perform pose estimation through joint optimization over spatial and temporal constraints with visual, inertial, and GNSS measurements, thereby improving global consistency via loop closures. Nevertheless, SLAM systems are susceptible to drift accumulation and environmental degeneracy, particularly in texture-less or geometrically sparse scenes.

In the past decade, deep neural network has been widely investigated for end-to-end pose estimation \cite{bib6, bib7}, which providing a promising solution for integrating multi-modal sensor observations. Despite their success, existing deep learning-based approaches still have several critical limitations. First, hierarchical fusion of heterogeneous sensor data remains challenging, as different modalities inherently capture distinct physical states. Camera expresses local motion information through pixel shifting, while IMU measures higher-order derivatives of high-frequency motion, and GNSS provides absolute positioning values. Existing methods often resort to shallow feature concatenations, failing to deeply align these representations and capture kinematic correlations. Second, temporal modeling and sensor asynchrony are critical difficult, effectively aligning and integrating multi-modal sensor data streams with inconsistent sampling frequencies and asynchronous characteristics within the network architecture remains a challenge which reduce the accuracy of pose estimation. Third, purely data-driven approaches easily violate physical constraints, pose estimation via the black-box networks— solely relying on data distributions without physical priors—tends to violate inherent constraints such as vehicle dynamics.

To address the above challenges, we propose a novel end-to-end deep neural network, DAP-Pose that fuse multi-modal sensors toward robust pose estimation. As illustrated in Fig. \ref{fig:DAP-Pose}, the network extracts modality-specific motion features from the three data modalities through independent encoders and performs pose prediction via three key modules. The main contributions are summarized as follows:

\begin{figure*}[t]  
\centering  
\includegraphics[width=\textwidth]{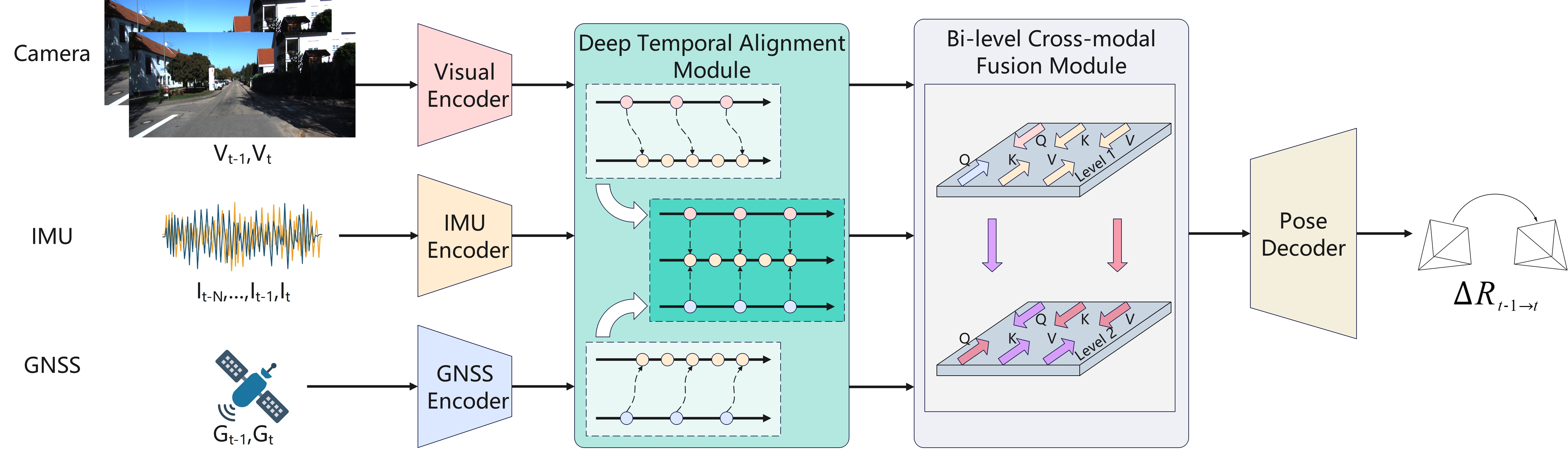}  
\caption{The overall architecture of DAP-Pose: the model takes measurements of multi-modal sensors, i.e., camera, IMU, and GNSS. The data streams sequentially pass through their respective feature encoders, the DTA module, the BCF module, and the Pose Decoder to estimate the relative pose between consecutive frames.}
\label{fig:DAP-Pose}
\end{figure*}

\begin{itemize}
\item A novel Bi-level Cross-modal Fusion (BCF) module is introduced to capture fine-grained kinematic interactions across heterogeneous sensors at both representation and motion levels.

\item A Deep Temporal Alignment (DTA) module is introduced to solve asynchronous problem of different sensors. The module explicitly learns to align asynchronous sensor streams in latent space, effectively mitigating temporal misalignment without manual interpolation or hardware-level synchronization.

\item Physics-aware pose modeling with motion-consistency constraints were imposed. By explicitly incorporating physical kinematic priors via $SO(3)$ manifold geometric constraints and GNSS-guided absolute metric scale, we strictly constrain the network's predictions within a valid rigid-body motion space.
\end{itemize}

\section{Related Works}

Traditional geometry-based multi-sensor fusion methods typically rely on algorithms of filtering \cite{bib8, bib9} or optimization \cite{bib5, bib10, bib11, bib12}.However, these methods heavily depend on hand-crafted features and complex fusion mechanisms, they are susceptible to critical failures in degraded environments \cite{bib13}.

To overcome the limitations, end-to-end deep learning models have been proposed to utilize neural networks to automatically learn and fuse complementary features. However, current end-to-end approaches are predominantly limited to bi-modal fusion. Existing network architectures, such as VINet \cite{bib6} and subsequent methods \cite{bib14, bib15, bib16, bib17, bib18}, typically relied on direct feature concatenation and utilized simple networks to weight and fuse the representations. However, these primitive strategies often suffer from weak feature representation and significant information loss. To improve cross-modal interactions, attention-based fusion networks \cite{bib19, bib20} and external memory attention-based fusion mechanism \cite{bib21} were proposed. Recently, transformer-based architectures have been introduced to effectively capture long-range temporal dependencies \cite{bib7, bib22}, and RWKV networks have also been explored for efficient multi-modal fusion \cite{bib23}. Furthermore, cross-attention networks \cite{bib24, bib25} have been introduced to enable more fine-grained interactions and fusion across different modalities.

Despite the advancements, end-to-end frameworks that simultaneously fuse camera, IMU, and GNSS remain scarce. Among existing attempts, some studies adopt a cascaded strategy, for instance, model adopts VO (Visual Odometry) via CNN-LSTM and subsequently fusing GNSS and INS \cite{bib26}. Other approaches \cite{bib27, bib28} employ neural networks but still rely on traditional filtering algorithms for the core fusion process. Consequently, we still lack of unified end-to-end networks capable of directly and jointly fusing data from visual sensor, inertial and GNSS measurements.

    Moreover, even hardware synchronization \cite{bib29} is a widely adopted solution to ensure temporal consistency, data alignment is still a costly endeavor and lacks scalability across diverse low-cost platforms. To address the issue, many software-based temporal calibration algorithms have been developed. Offline approaches, such as Kalibr \cite{bib30} and curve-alignment methods \cite{bib31}, achieve precise temporal calibration prior to operation. For online calibration, current methods typically incorporate the temporal offset as an optimizable state variable within the filtering or optimization process \cite{bib32}.

To date, deep learning based explicit temporal alignment still have a notable gap for practical application. Although various data-driven methods were applied for  LiDAR-Camera synchronization \cite{bib33}, deep temporal alignment among visual, inertial, and GNSS sensors is largely underexplored. Although TON-VIO \cite{bib34} proposes a lightweight online time offset modeling network for VIO (Visual-Inertial Odometry) systems, the vast majority of existing end-to-end multi-modal networks operate under the assumption of perfect synchronization, solely relying on the network's implicit robustness to tolerate minor temporal offsets. Consequently, they are prone to severe performance degradation under significant sensor asynchrony. This highlights the urgent need for explicit temporal alignment mechanisms in deep multi-modal fusion frameworks.

\section{Methodology}

\subsection{Problem Formulation} 

DAP-Pose aims to achieve robust localization and orientation estimation. To address the challenges, the model estimates 6-DoF relative poses between consecutive visual frames by fusing multi-level semantic and geometric representations. Let $V_{t-1}$ and $V_{t}$ denote two consecutive image timestamps, we denote all captured high-frequency IMU measurements and GNSS coordinate measurements as $I_{t-1:t}$ and $G_{t-1:t}$, respectively between the image timestamps. Given these multi-modal inputs, the network aims to continuously output the relative pose transformation $\Delta T_{t-1 \to t} \in \text{SE(3)}$, which consists of a translation component $\Delta p \in \mathbb{R}^3$ and a rotation component $\Delta R \in \text{SO(3)}$.

\subsection{Sensor Data Encoders}

To bridge the significant representational differences and inconsistent sampling frequencies among heterogeneous sensors, we design three modality-specific encoders. These encoders aim to uniformly map physical observations at different frequencies within the same time window into distinct semantic and geometric features, denoted as $F_t^V$, $F_t^I$, and $F_t^G$, respectively.

Visual Encoder: To implicitly encode the images with rich structural semantics information, a FlowNetSimple-based architecture is introduce to extract optical flow features between consecutive image frames, that explicitly captures the semantic representations of local geometric displacements.

Inertial Encoder: As higher-order derivatives of motion, IMU measurements reflect high-frequency kinematic states. We employ a 1D CNN to process IMU data sequences, leveraging its local temporal receptive field to efficiently extract kinematic features, thereby providing continuous and dense ego-motion priors for the network.

GNSS Encoder: Unlike relative measurements of images and IMU measurements, GNSS provides drift-free global constraints. To ensure spatial translation invariance and avoid overfitting caused by absolute coordinates, the transformer-based encoder takes 3D coordinate differences and their Euclidean norm as inputs. Crucially, the Euclidean norm serves as an explicit physical prior, injecting an absolute metric scale and motion consistency constraint into the network. This achieves global geometric consistency and resolves the inherent local scale ambiguity in visual-inertial systems.

\subsection{Deep Temporal Alignment}
\begin{figure}[t]  
\centering  
\includegraphics[width=\linewidth]{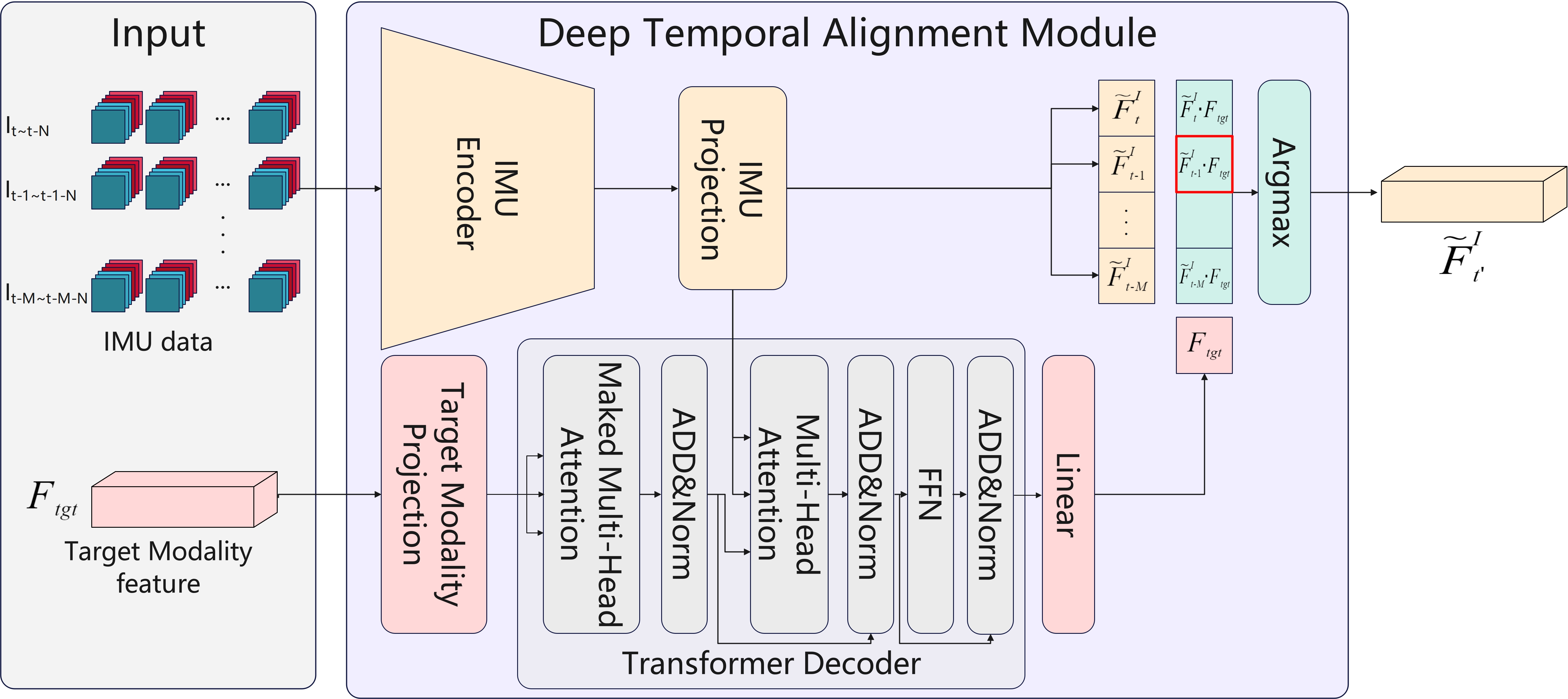}  
\caption{Network structure of DTA: the module explicitly aligns asynchronous sensor streams by modeling the cross-modal temporal context via cross-attention, and extracts the optimally synchronized inertial feature based on the maximum cosine similarity.}
\label{fig:DTA}
\end{figure}
Existing end-to-end multi-modal networks typically assume multi-modal inputs are perfectly synchronized. However, sensor asynchrony inevitably occurs in real-world systems. While feature encoders address inconsistent sampling frequencies across sensors, the extracted features often exhibit temporal offsets on the absolute timeline. Directly fusing these misaligned asynchronous features causes conflicting physical motion cues in the network. To address the issue, we propose the DTA module. By utilizing the sparse visual or GNSS features as alignment anchors, this module adopts a sliding windows within dense, high-frequency IMU sequences for local search, and explicitly aligns the asynchronous data streams in a latent feature space. This enables coherent motion modeling without requiring strict hardware synchronization.

As illustrated in Fig. \ref{fig:DTA}, the DTA module processes the inputs in a pairwise manner (e.g., Visual-Inertial or GNSS-Inertial). Within each pair, temporally sparse visual or GNSS features serve as the single-frame target feature $F_{tgt}$. Simultaneously, a sliding window with a stride of 1 is applied to the continuous high-frequency IMU data stream, generating a reference feature sequence $\mathcal{F}_{ref}=\{F_{t-M}^I, \dots, F_{t}^I\}$ covering various potential temporal offsets via the inertial encoder, where $M$ denotes the length of the reference feature sequence in one alignment process.

First, $F_{tgt}$ and $\mathcal{F}_{ref}$ are projected into a shared latent space to extract cross-modal shared kinematic features, yielding $\widetilde{F}_{tgt}$ and $\widetilde{\mathcal{F}}_{ref}$. Subsequently, we employ a Transformer decoder for deep cross-modal temporal modeling, where the projected target features serve as the Query, and the reference feature sequence acts as the Key and Value. The decoder's output feature $F_{out}$ captures the temporal motion context most relevant to the target feature state:
$$F_{out} = \text{TransformerDecoder}(\widetilde{F}_{tgt}, \widetilde{\mathcal{F}}_{ref}, \widetilde{\mathcal{F}}_{ref}) \eqno{(1)}$$

Finally, by computing the cosine similarity between $F_{out}$ and each projected inertial feature ${\widetilde{F}}_i^I \in \widetilde{\mathcal{F}}_{ref}$ in the reference sequence, the network identifies the index that yields the maximum similarity, and regards the corresponding temporal offset as the optimal synchronized temporal offset for the target feature:
$$\text{Similarity}_{i} = \frac{F_{out} \cdot {\widetilde{F}}_i^I}{\|F_{out}\| \|{\widetilde{F}}_i^I\|} \eqno{(2)}$$

To address the non-differentiability of the $argmax$ operation during temporal offset selection, we replace the hard matching process using a $Gumbel-Softmax$ approximation. Specifically, a soft alignment distribution is constructed over all candidate temporal offsets:
$$P_{i} = Gumbel-Softmax(
\frac{\text{Similarity}_{i}}{\tau}) \eqno{(3)}$$
where $\tau$ is a temperature parameter that controls the sharpness of the distribution. During training, this continuous relaxation enables gradient backpropagation through the alignment module, allowing the DTA module to be jointly optimized with the pose regression objective in an end-to-end manner.

The aligned feature representation is then computed as a weighted aggregation:
$$F_{align} = \sum_{i=t-M}^{t} P_i \cdot F_i^I \eqno{(4)}$$

During inference, we adopt a deterministic strategy via hard selection of the index with maximum probability, which produces the final estimated temporal offset.

This design ensures differentiability during training while preserving explicit hard alignment during deployment.

The temporal modeling mechanism, based on explicit metric computation in feature space, adaptively achieves the alignment of asynchronous multi-modal signals, thereby fundamentally breaking the reliance of conventional networks on hardware synchronization.

\subsection{Bi-level Cross-modal Fusion}
\begin{figure}[t]  
\centering  
\includegraphics[width=\linewidth]{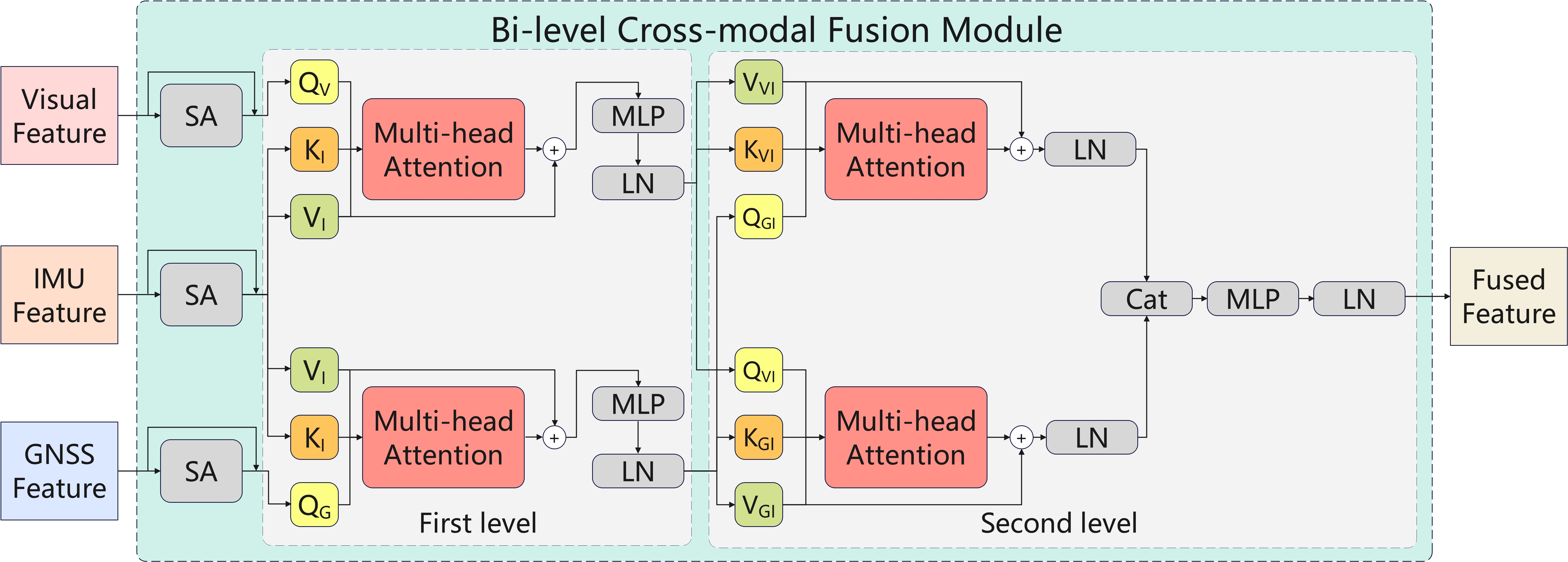}  
\caption{Architecture of BCF module, where the left and right components represent the first and second levels, respectively: the first level enforces motion consistency using Visual and GNSS features to independently query the continuous IMU representations, and the second level achieves deep semantic-geometric fusion via bidirectional cross-attention between the two fused branches. (SA: Self-Attention; LN: Layer Normalization; Cat: Concatenation; MLP: Multi-Layer Perceptron).}
\label{fig:BCF}
\end{figure}
Generally, multi-modal sensor data exhibit significant hierarchical differences in  physical representations: visual features represent local structural semantics and relative spatial displacements, inertial features encode continuous high-frequency kinematic states, while GNSS measurements provide global geometric constraints with absolute metric scale. Existing end-to-end fusion methods often rely on direct feature concatenation \cite{bib6, bib14, bib15, bib16, bib17, bib18} or simple attention mechanisms \cite{bib7, bib19, bib20, bib21, bib22}. Such shallow fusion strategies struggle to handle the deep nonlinear correlations among heterogeneous modalities and overlook the inherent measurement properties of the observations. To address the hierarchical representational differences and achieve deep multi-modal fusion, we propose BCF module, which aims to model the fine-grained interactions among heterogeneous sensors at both representation and motion levels, thereby achieving the multi-level fusion of data semantic information and geometric localization information.

As illustrated in Fig. \ref{fig:BCF}, let $\mathcal{F}_V=\{F_{t-N-1}^V, \dots, F_t^V\}$, $\mathcal{F}_I=\{F_{t-N-1}^I, \dots, F_t^I\}$, and $\mathcal{F}_G=\{F_{t-N-1}^G, \dots, F_t^G\}$ denote the aligned feature sequences, where $N$ represents the sequence length. These sequences are first passed through respective self-attention layers to obtain the enhanced intra-modal motion features $\mathcal{F}'_V$, $\mathcal{F}'_I$, and $\mathcal{F}'_G$. Subsequently, these features perform deep fusion sequentially through motion-level and representation-level interactions:

The first level is designed for cross-modal motion consistency modeling, aims to resolve the complementarity and alignment issues between discrete observations and continuous motion history, the parallel Visual-Inertial and GNSS-Inertial cross-attention units are introduced. Given the physical observation characteristics of sensors, visual and GNSS features represent discrete states at the boundaries of a temporal window, while inertial features encapsulate the dense, continuous motion trajectory prior within that window. Therefore, in these two attention units, the enhanced inertial feature is set as the key and value to serve as an information anchor, and utilize the enhanced visual features and GNSS features as the query respectively. This explicit asymmetric query mechanism enables the discrete semantic and geometric observation states to extract highly relevant local kinematic contexts from the continuous motion history. Through interaction in this layer, the network effectively filters out ego-motion-irrelevant semantic disturbances in the visual modality (e.g., dynamic obstacles, abrupt illumination changes) and local noise in the GNSS modality, generating bi-modal features $F_{VI}$ and $F_{GI}$ with strict kinematic consistency:
$$F_{VI} = \text{LN}(\text{MLP}(\text{CrossAttn}(\mathcal{F}'_V, \mathcal{F}'_I, \mathcal{F}'_I))) \eqno{(5)}$$
$$F_{GI} = \text{LN}(\text{MLP}(\text{CrossAttn}(\mathcal{F}'_G, \mathcal{F}'_I, \mathcal{F}'_I))) \eqno{(6)}$$

The second level is the semantic-geometric representation fusion. The outputs from the first level exhibit significant complementarity at the representation level: $F_{VI}$ contains fine-grained local semantics and motion states, whereas $F_{GI}$ carries global absolute scale constraint and geometric localization information. To achieve deep information interaction at the representation level, this layer applies bidirectional cross-attention to both. $F_{GI}$ queries $F_{VI}$ to obtain local structured semantic details, while $F_{VI}$ queries $F_{GI}$ to introduce global absolute metric scales and motion consistency constraints, thereby suppressing drift. Finally, the outputs of the two branches ($F_{GIV}$ and $F_{VIG}$)from the bidirectional interaction are concatenated and nonlinearly mapped by a Multi-Layer Perceptron (MLP), generating a joint motion representation $F_{fuse}$ that possesses both strong geometric consistency and local semantic awareness:
$$F_{fuse} = \text{LN}(\text{MLP}([F_{GIV},F_{VIG}]))) \eqno{(7)}$$

This hierarchical fusion network successfully unifies physics-awareness with data-driven learning, providing highly robust feature representations for subsequent high-precision pose regression.

\subsection{Physics-Aware Pose Modeling}

To derive the 6-DoF relative pose transformation, the fused multi-modal representation is fed into a lightweight CNN-based pose decoder, as the decoder avoids the sequential computational inefficiency of LSTM-based decoders and the parameter redundancy of MLP-based decoders. To align with the latent space of the pre-trained encoders, the decoder outputs the relative translation $\Delta \hat{p} \in \mathbb{R}^3$ and the relative rotation $\Delta \hat{\theta} \in \mathbb{R}^3$, utilizing Euler angles as an intermediate parametric representation. Subsequently, the system analytically maps these Euler angles into a rotation matrix $\Delta \hat{R} \in SO(3)$. The proposed network is trained end-to-end in a supervised manner. 

Traditional data-driven networks directly apply mean squared error for predicting Euler angle. However, limited by the non-uniqueness and gimbal lock problems of Euler angles, the numerical difference-based loss formulation usually fail to account for the rigorous topological structure of 3D rotations, making it difficult for the network outputs to satisfy rigid-body motion consistency. To address this issue, we introduce geometric priors to strictly constrain rotation errors within the $SO(3)$ manifold. The rotation loss $\mathcal{L}_{rot}$ consists of two physically meaningful components. First, following the work of \cite{bib7}, we similarly employ the Regularized Projective Manifold Gradient (RPMG) layer  \cite{bib36} to compute the manifold loss $\mathcal{L}_{RPMG}$. Second, we analytically convert the predicted Euler angles into a rotation matrix and utilize the Frobenius norm to penalize the relative error matrix between it and the ground truth matrix $\Delta R_{gt}$. This operation transforms the loss objective from a mere numerical difference in Euler angles to a genuine 3D geometric orientation difference. Essentially, it computes the manifold chordal distance in the $SO(3)$, explicitly measuring the geometric error within the rigid-body rotation manifold:

$$\mathcal{L}_{rot} = \mathcal{L}_{RPMG} + \alpha \|\Delta R \Delta R_{gt}^T - I\|_F^2 \eqno{(8)}$$
where $\alpha$ is a scaling factor and I is the identity matrix. This design, which guides network optimization via physical geometric structures, not only avoids topological singularities inherent in purely data-driven regression but also yields smoother backpropagation gradients, substantially enhancing the training stability of the network in complex motion scenarios.

For the translation component, leveraging the absolute metric scale provided by GNSS feature injection and the motion consistency soft constraints from the BCF module, we compute the translation loss $\mathcal{L}_{trans}$ using the $\mathcal{L}_{1}$ distance:
$$\mathcal{L}_{trans} = \|\Delta p - \Delta p_{gt}\|_1 \eqno{(9)}$$

Furthermore, to explicitly supervise the DTA module, we introduce a Temporal Alignment Loss $\mathcal{L}_{align}$ formulated as a triplet margin objective. By defining the precisely time-aligned inertial feature as the positive sample and temporally unaligned inertial features as negative samples, the loss is defined as:
$$\mathcal{L}_{align} = \sum_{j=1}^{M} \max(\text{Similarity}_j - \text{Similarity}_p + m, 0) \eqno{(10)}$$
where $p$ denotes the positive sample, $M$ is the number of unaligned inertial features within the sliding window, and m is a predefined margin.

Finally, the overall physics-aware optimization objective $L_{total}$ of the end-to-end network is a weighted combination of the aforementioned terms:
$$\mathcal{L}_{total} = \mathcal{L}_{trans} + \lambda_1 \mathcal{L}_{rot} + \lambda_2 \mathcal{L}_{align} \eqno{(11)}$$
where $\lambda_1$ and $\lambda_2$ are balancing weights.

\section{Experiments}

\subsection{Datasets}

We evaluate the proposed network on widely used KITTI dataset \cite{bib38}. We adopt the left-camera, IMU, and GNSS data. The raw IMU data is captured at 100 Hz, while the left-camera images, GNSS data, and ground truth trajectories are recorded at 10 Hz. Following the settings of works such as \cite{bib7} and \cite{bib14}, we use sequences 00, 01, 02, 04, 06, 08, and 09 for training, and sequences 05, 07, and 10 for testing.

\subsection{Training and Evaluation}

The proposed network is implemented with the PyTorch framework. The pre-trained visual and inertial encoders \cite{bib14} are adopted for feature extraction. During training, all input images are resized to a resolution of $512×256$. The input sequence length is set to 11. In the feature extraction stages, the feature dimensions for the visual, inertial, and GNSS modalities are configured to 512, 256, and 128, respectively. The final multi-modal fused feature dimension is set to 512. The network is trained for 200 epochs with a batch size of 128. We employ a cosine annealing learning rate scheduler, restarting every 25 epochs, with an initial learning rate of $1×10^{-4}$.

We employ $t_{rel}$ and $r_{rel}$ as the evaluation metrics for our experiments. Specifically, $t_{rel}$(\%) measuring the average translation error percentage over sub-sequence lengths ranging from 100 to 800 meters. Similarly, $r_{rel}$($^{\circ}$/100m) evaluating the average rotation error over the same sub-sequence lengths.

\subsection{Results}

\begin{table*}[t]
    \centering
    \caption{Quantitative comparison of methods upon the KITTI dataset: the best and second-best results are highlighted in bold and underline, respectively.}
    \label{tab:kitti_comparison}
    \begin{tabular}{l c cc cc cc cc}
        \toprule
        \multirow{2}{*}{Model} & \multirow{2}{*}{Type} & \multicolumn{2}{c}{Seq.05} & \multicolumn{2}{c}{Seq.07} & \multicolumn{2}{c}{Seq.10} & \multicolumn{2}{c}{Average} \\
        \cmidrule(lr){3-4} \cmidrule(lr){5-6} \cmidrule(lr){7-8} \cmidrule(lr){9-10}
        & & $t_{\mathrm{rel}}$(\%) & $r_{\mathrm{rel}}(^\circ)$ & $t_{\mathrm{rel}}$(\%) & $r_{\mathrm{rel}}(^\circ)$ & $t_{\mathrm{rel}}$(\%) & $r_{\mathrm{rel}}(^\circ)$ & $t_{\mathrm{rel}}$(\%) & $r_{\mathrm{rel}}(^\circ)$ \\
        \midrule
        VINS-Mono \cite{bib5}   & Geo.      & 11.6  & 1.26 & 10.0  & 1.72 & 16.5  & 2.34 & 12.7  & 1.77 \\
        ROVIO \cite{bib8}       & Geo.      & 3.21  & 1.22 & 2.97  & 1.38 & 3.20  & 1.33 & 3.13  & 1.31 \\
        \midrule
        VIOLearner \cite{bib37} & Self-Sup. & 3.00  & 1.40 & 3.60  & 2.06 & 2.04  & 1.37 & 2.88  & 1.61 \\
        DeepVIO \cite{bib15}    & Self-Sup. & 2.86  & 2.32 & 2.71  & 1.66 & \textbf{0.85} & 1.03 & 2.14  & 1.67 \\
        BotVIO \cite{bib25}     & Self-Sup. & -     & -    & -     & -    & 6.56  & \textbf{0.08} & -    & -    \\
        \midrule
        ATVIO \cite{bib20}      & Sup.      & 4.93  & 2.40 & 3.78  & 2.59 & 5.71  & 2.96 & 4.81  & 2.65 \\
        Hard Fusion \cite{bib18}& Sup.      & 4.11  & 1.49 & 3.44  & 1.86 & 1.51  & 0.91 & 3.02  & 1.42 \\
        VIOFormer \cite{bib22}  & Sup.      & -     & -    & -     & -    & 2.34  & 0.74 & -     & -    \\
        Yang et al. \cite{bib14}& Sup.      & 2.04 & \underline{0.76} & \underline{1.75} & 0.71 & 3.58  & 1.12 & 2.46  & 0.86 \\
        VIFT \cite{bib7}        & Sup.      & 2.47  & 0.88 & 1.77  & 1.06 & 1.73  & 0.63 & 1.99 & 0.85 \\
        RWKV-VIO \cite{bib23}   & Sup.      & \underline{2.03}  & 1.00 & 2.73  & 1.79 & 2.10  & 0.99 & 2.29  & 1.26 \\
        \midrule
        Ours-VI            & Sup.      & 2.48 & 0.80 & 1.86 & \underline{0.46} & 1.36 & \underline{0.37} & \underline{1.90} & \underline{0.54} \\
        Ours            & Sup.      & \textbf{1.60} & \textbf{0.53} & \textbf{1.16} & \textbf{0.44} & \underline{1.16} & 0.41 & \textbf{1.31} & \textbf{0.46} \\
        \bottomrule
    \end{tabular}
\end{table*}

Since the KITTI dataset is inherently time-synchronized, the DTA module is not applied in the comparative experiments of this section. This ensures a fair comparison of the intrinsic multi-modal fusion performance of our network against other state-of-the-art methods.

To evaluate performance of the proposed network, we take the representative state-of-the-art VIO algorithms as baseline, given the current scarcity of end-to-end learning methods that fuse visual, inertial, and GNSS modalities. This comparison aims to comprehensively evaluate the effectiveness of the proposed multi-modal fusion network in achieving highly accurate and robust pose estimation. Specifically, we compare our approach against representative geometry-based methods \cite{bib5, bib8}, self-supervised learning methods \cite{bib15, bib25, bib37}, and supervised method \cite{bib7, bib14, bib18, bib20, bib22}. Among them, \cite{bib22, bib25} is trained on sequences 00-08 and tested on sequences 09-10. The training and testing data configurations for all other learning-based baselines are identical to our approach.

As summarized in Table \ref{tab:kitti_comparison}, the proposed method achieves state-of-the-art overall performance, yielding the lowest average $t_{rel}$ of 1.31\% and $r_{rel}$ of 0.46$^{\circ}$. Compared to the highly competitive baseline VIFT \cite{bib7}, our method reduces these average errors by approximately 34.17\% and 45.88\%, respectively. This substantial margin clearly validates the effectiveness of our multi-modal fusion network.

Specifically, on sequences 05 and 07, our approach consistently ranks first. Traditional geometry-based methods like VINS-Mono \cite{bib5} suffer from severe scale drift in these long-distance scenarios. While learning-based VIO methods mitigate this, our network further reduces the $t_{rel}$ to 1.60\% and 1.16\%, demonstrating that integrating the GNSS modality successfully resolves local scale ambiguity. Additionally, the outstanding rotational accuracy confirms the necessity of the imposed $SO(3)$ manifold constraint.

We noted that on sequence 10, DeepVIO \cite{bib15} and BotVIO \cite{bib25} achieve the lowest $t_{rel}$ and $r_{rel}$, respectively. However, DeepVIO exhibits noticeable performance fluctuations across other trajectories, and BotVIO suffers from a severe translation error with a $t_{rel}$ of 6.56\%. In contrast, our method maintains the second-best accuracy on Seq. 10 while ensuring stable reliability and exceptional generalizability across all evaluated scenarios.

\begin{figure*}[t] 
    \centering
    \begin{subfigure}{0.32\textwidth}
        \centering
        \includegraphics[width=\linewidth]{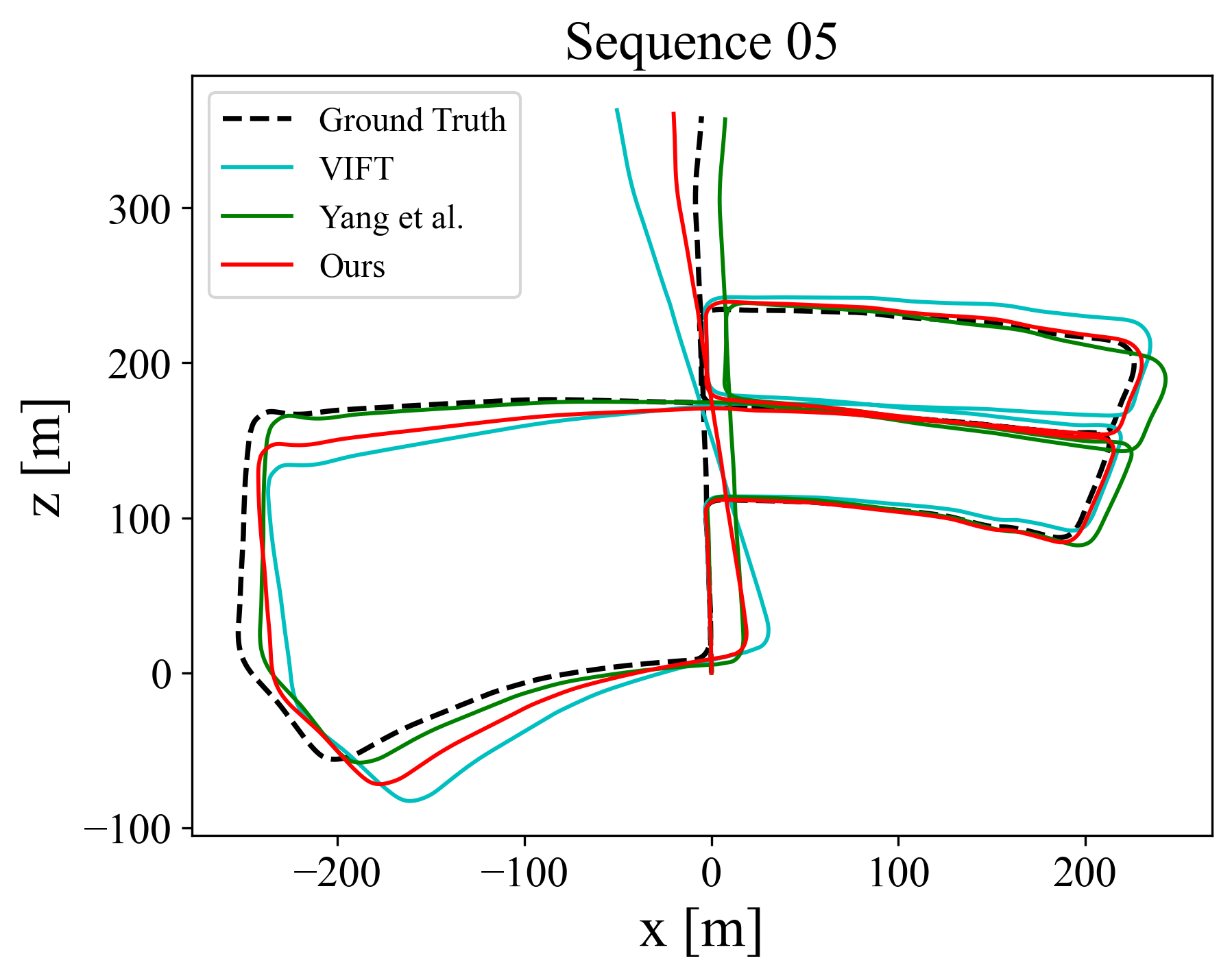} 
        \label{fig:sub_a}
    \end{subfigure}
    \hfill 
    \begin{subfigure}{0.32\textwidth}
        \centering
        \includegraphics[width=\linewidth]{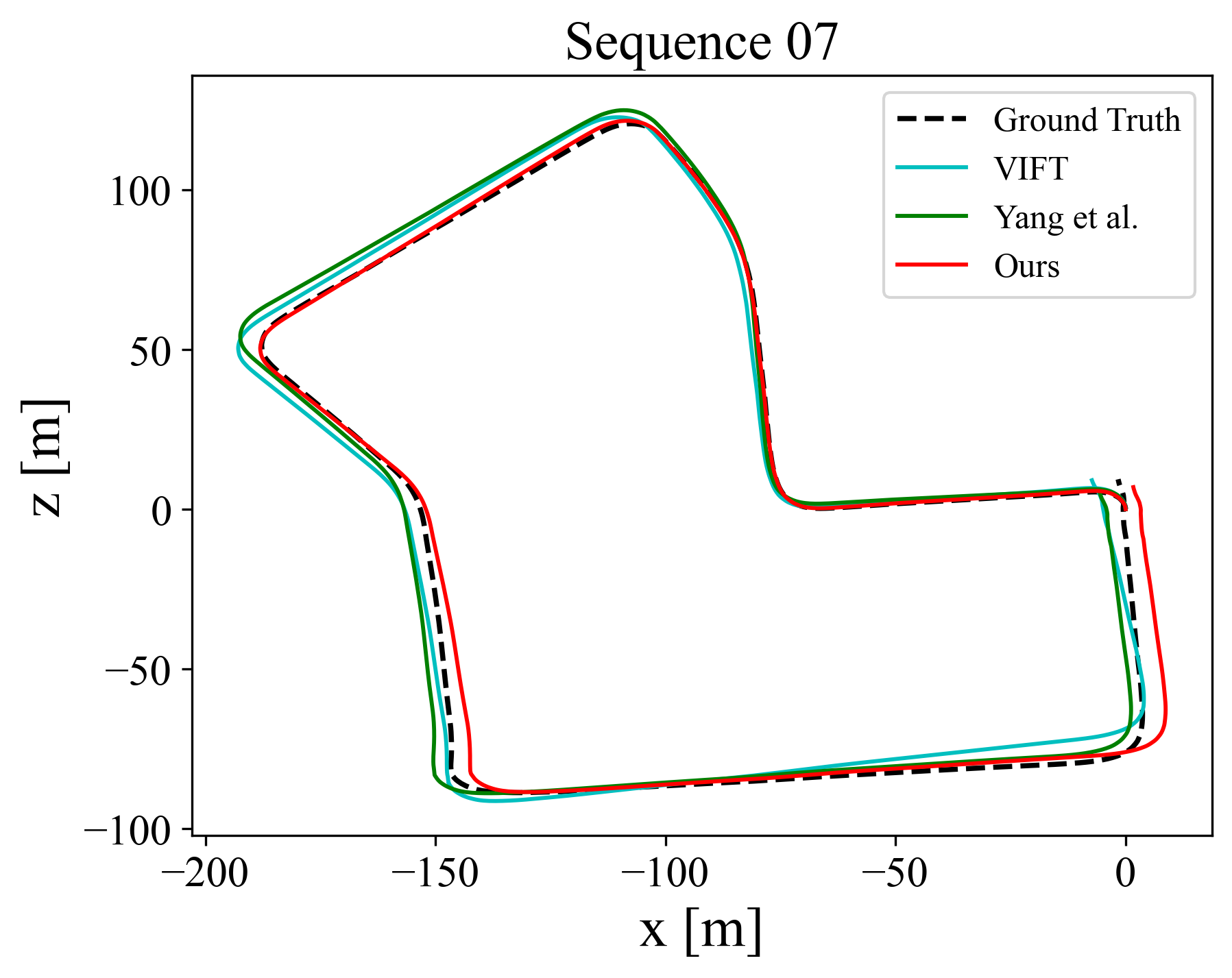} 
        \label{fig:sub_b}
    \end{subfigure}
    \hfill
    \begin{subfigure}{0.32\textwidth}
        \centering
        \includegraphics[width=\linewidth]{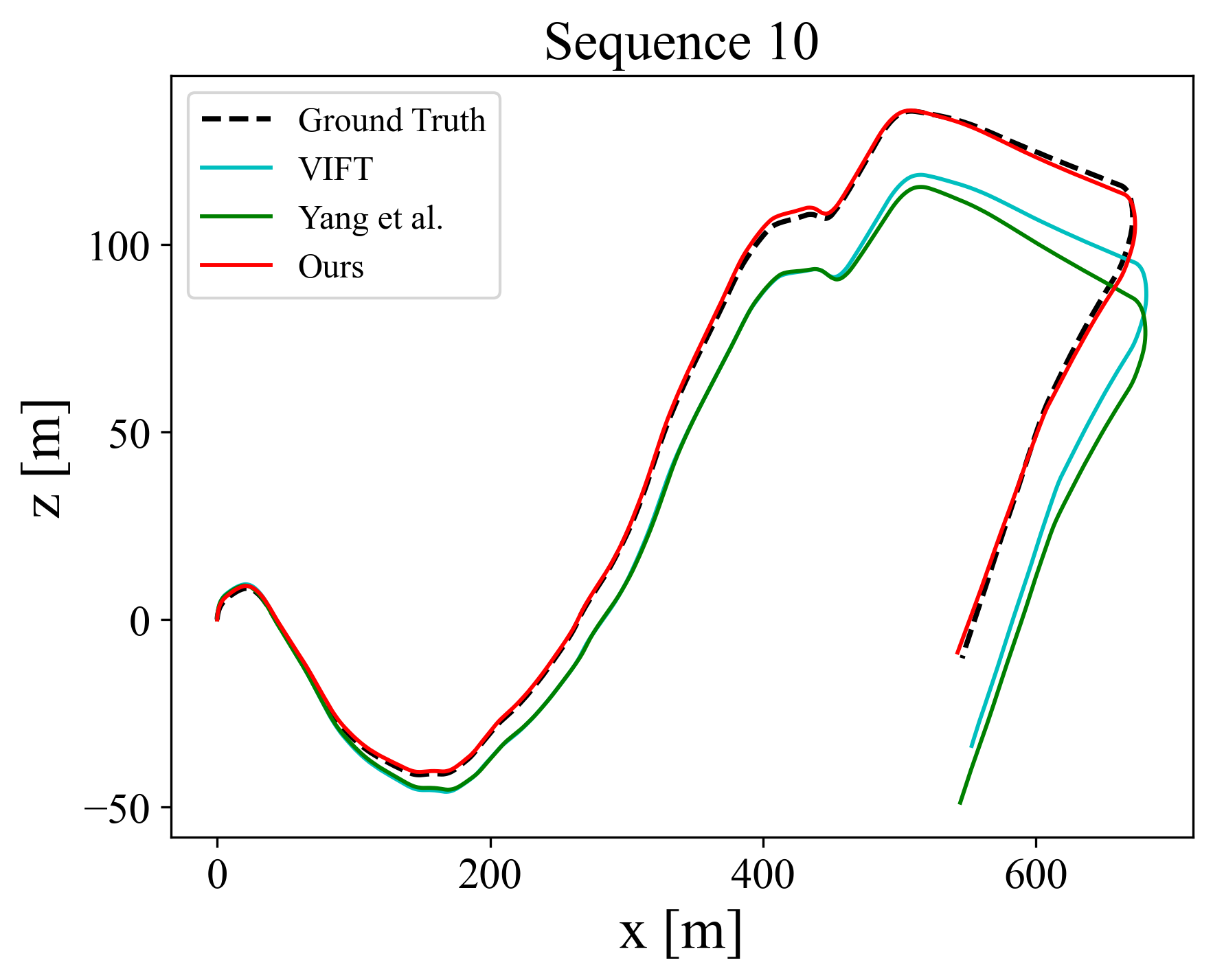} 
        \label{fig:sub_c}
    \end{subfigure}
    
    \caption{Trajectory comparison on the KITTI dataset. On sequences 07, 10, and the first half of sequence 05, the proposed method aligns more closely with the ground truth.}
    \label{fig:traj}
\end{figure*}

Furthermore, the qualitative trajectory visualization in Fig. \ref{fig:traj} intuitively confirms that our method aligns most closely with the ground truth compared to other competitive baselines.

It is worth noting that the comparisons in Table I involve different sensor configurations, where our full model additionally incorporates GNSS measurements, while most existing baselines rely only on visual-inertial inputs. To further evaluate the effectiveness of the proposed architecture independent of GNSS measurements, we construct an Ours-VI variant by replacing the GNSS branch with a visual branch in the BCF module while keeping the overall network architecture unchanged.

As shown in Table \ref{tab:kitti_comparison}, Ours-VI outperforms existing visual-inertial methods, achieving an average translation error of 1.90\% and rotation error of 0.54$^{\circ}$. This result demonstrates the effectiveness of the proposed hierarchical fusion architecture and physics-aware pose modeling even without GNSS measurements.

Furthermore, compared with Ours-VI, the full model further reduces the average translation error from 1.90\% to 1.31\%, while also improving the rotation accuracy from 0.54$^{\circ}$ to 0.46$^{\circ}$. This improvement demonstrates that GNSS provides global metric constraints, which primarily benefit translation estimation by constraining scale ambiguity and drift. Meanwhile, the additional modality also contributes to more consistent motion representation, leading to a slight improvement in rotation estimation.

\subsection{Evaluation of Deep Temporal Alignment}

\begin{table*}[t]
    \centering
    \caption{
    Pose estimation accuracy under artificial IMU delays. Bold indicates the best results in each block.}
    \label{tab:imu_delays}
    \begin{tabular}{c c cc cc cc cc}
        \toprule
        \multirow{2}{*}{Delay (s)} & \multirow{2}{*}{Model} & \multicolumn{2}{c}{Seq.05} & \multicolumn{2}{c}{Seq.07} & \multicolumn{2}{c}{Seq.10} & \multicolumn{2}{c}{Average} \\
        \cmidrule(lr){3-4} \cmidrule(lr){5-6} \cmidrule(lr){7-8} \cmidrule(lr){9-10}
        & & $t_{\mathrm{rel}}$(\%) & $r_{\mathrm{rel}}(^\circ)$ & $t_{\mathrm{rel}}$(\%) & $r_{\mathrm{rel}}(^\circ)$ & $t_{\mathrm{rel}}$(\%) & $r_{\mathrm{rel}}(^\circ)$ & $t_{\mathrm{rel}}$(\%) & $r_{\mathrm{rel}}(^\circ)$ \\
        \midrule
        0   & Ours w/o DTA & \textbf{1.60} & \textbf{0.53} & \textbf{1.16} & \textbf{0.44} & \textbf{1.16} & \textbf{0.41} & \textbf{1.31} & \textbf{0.46} \\
        \midrule
        0.1 & VIFT         & 2.58 & 0.96 & 2.10 & 1.24 & 2.21 & 0.80 & 2.30 & 1.00 \\
        0.1 & Ours w/o DTA & 2.03 & 0.78 & 1.77 & 0.87 & 1.63 & 0.59 & 1.81 & 0.75 \\
        0.1 & Ours w/ DTA  & \textbf{1.80} & \textbf{0.65} & \textbf{1.27} & \textbf{0.55} & \textbf{1.09} & \textbf{0.40} & \textbf{1.39} & \textbf{0.53} \\
        \midrule
        0.2 & VIFT         & 2.94 & 1.12 & 2.84 & 1.56 & 2.97 & 1.09 & 2.92 & 1.26 \\
        0.2 & Ours w/o DTA & 2.47 & 0.96 & 2.72 & 1.28 & 2.45 & 0.92 & 2.55 & 1.05 \\
        0.2 & Ours w/ DTA  & \textbf{1.84} & \textbf{0.68} & \textbf{1.16} & \textbf{0.57} & \textbf{1.11} & \textbf{0.40} & \textbf{1.37} & \textbf{0.55} \\
        \bottomrule
    \end{tabular}
\end{table*}

Since the original KITTI dataset is hardware-synchronized, we artificially inject fixed delays of 0.1 and 0.2 seconds into the IMU data streams to evaluate the Deep Temporal Alignment module. Under these asynchronous setups, we assess the pose estimation accuracy of our network with and without this module, comparing the results against the highly competitive baseline VIFT.

As shown in Table \ref{tab:imu_delays}, artificial IMU delays degrade the pose estimation accuracy of all methods. Notably, even without the DTA module, our network consistently outperforms the VIFT baseline under identical uncalibrated conditions.

When the proposed DTA module is employed, its explicit feature alignment, coupled with the network's inherent tolerance to temporal offsets, effectively compensates for severe misalignments. With the configuration of 0.1s and 0.2s delays, $t_{rel}$ recovers to 1.39\% and 1.37\%, closely approaching the zero-delay baseline. Meanwhile, $r_{rel}$ is restored to 0.53$^{\circ}$ and 0.55$^{\circ}$, with some results marginally better than the baseline. This is primarily because on sequences 07 and 10, minor temporal shifts during alignment process coincidentally offset the inherent regression bias. Overall, the DTA module effectively resolves sensor asynchrony.

\section{Conclusion}

In this paper we propose DAP-Pose, a unified end-to-end neural network for robust Visual-Inertial-GNSS pose estimation under complex environments with asynchronous sensor data. By jointly addressing cross-modal representation learning, temporal misalignment, and physics-aware motion modeling within a neural network, DAP-Pose provides a principled solution to long-standing challenges in heterogeneous sensor fusion for pose estimation.

Specifically, a novel BCF module is introduced that enables hierarchical interactions across multiple modalities at both feature and motion levels, effectively leveraging the complementary position and orientation information. To overcome the issues caused by asynchronous measurements, the DTA module is proposed to learn latent temporal correspondence directly from data, eliminating the reliance on strict hardware synchronization or heuristic interpolation. Furthermore, by embedding $SO3$ manifold geometric constraints and GNSS-guided absolute metric scale into the learning process, the model incorporates physics-aware motion-consistency priors that explicitly enforce rigid-body kinematics, improving geometric reliability and long-horizon stability.

Extensive experiments on the KITTI datasets suggest that DAP-Pose achieves SOTA accuracy while exhibiting remarkable robustness to temporal offsets. The results validate not only the effectiveness of each proposed component but also the advantage of jointly modeling cross-modal interaction, temporal alignment, and physical consistency in a unified framework.

Future work will extend DAP-Pose toward broader heterogeneous sensor configurations, including LiDAR, and event cameras, to explore adaptive uncertainty modeling for enhanced reliability in safety-critical autonomous systems. We believe the proposed framework offers a scalable foundation for next-generation robust multi-modal navigation on mobile robotic platforms operating in real-world environments.





\bibliographystyle{IEEEtran}
\bibliography{IEEEabrv,reference}

\end{document}